\pdfoutput=1

\documentclass[11pt]{article}

\usepackage[]{ACL2023}

\usepackage{times}
\usepackage{latexsym}
\usepackage{subfig} 
\usepackage[T1]{fontenc}

\usepackage[utf8]{inputenc}

\usepackage{microtype}

\usepackage{inconsolata}
\usepackage{amsmath}
\usepackage{makecell}
\usepackage{multirow}
\usepackage{booktabs}
\usepackage{graphicx}
\usepackage{algorithm}
\usepackage{algorithmicx}
\usepackage{algpseudocode}
\title{Benchmarking Knowledge Boundary for Large Language Models: A Different Perspective on Model Evaluation}

\author{Xunjian Yin\footnotemark[1],~~Xu Zhang\footnotemark[1],~~Jie Ruan \and  Xiaojun Wan \\
        Wangxuan Institute of Computer Technology, Peking University \\
        \texttt{\{xjyin, zhangxu, wanxiaojun\}@pku.edu.cn} \\
        \texttt{ruanjie@stu.pku.edu.cn}
        }
        
\begin{document}
\maketitle

\renewcommand{\thefootnote}{\fnsymbol{footnote}}
\footnotetext[1]{These authors contributed equally to this work.}
\renewcommand*{\thefootnote}{\arabic{footnote}}

\begin{abstract}
In recent years, substantial advancements have been made in the development of large language models, achieving remarkable performance across diverse tasks.
To evaluate the knowledge ability of language models, previous studies have proposed lots of benchmarks based on question-answering pairs.
We argue that it is not reliable and comprehensive to evaluate language models with a fixed question or limited paraphrases as the query, since language models are sensitive to prompt.
Therefore, we introduce a novel concept named knowledge boundary to encompass both prompt-agnostic and prompt-sensitive knowledge within language models.
Knowledge boundary avoids prompt sensitivity in language model evaluations, rendering them more dependable and robust.
To explore the knowledge boundary for a given model, we propose a projected gradient descent method with semantic constraints, a new algorithm designed to identify the optimal prompt for each piece of knowledge.
Experiments demonstrate a superior performance of our algorithm in computing the knowledge boundary compared to existing methods.
Furthermore, we evaluate the ability of multiple language models in several domains with knowledge boundary.
\end{abstract}

\section{Introduction}
Recently, large language models (LLMs) have made significant advancements in a variety of tasks \citep{brown2020language,thoppilan2022lamda,bubeck2023sparks}.
In order to gain deeper insights into the knowledge capabilities of different LLMs to help select appropriate LLM in practice, numerous studies have proposed various benchmarks for LLM evaluation \citep{guo2023close,zhong2023agieval}.
The majority of previous research on model evaluation constructs a test dataset sourced from standardized examinations, such as college entrance exams and law school admission tests \citep{hendrycks2021measuring}. 
Subsequently, the questions are fed to LLMs as prompts, eliciting responses that are then scored for evaluation \cite{yu2023kola, zhang2023benchmarking}.

However, each piece of knowledge embodies \textbf{abstract concept} that can be expressed in a nearly infinite number of \textbf{textual forms} \citep{knowledgearck}.
When evaluating a specific piece of knowledge, existing work only evaluated LLMs with one or several textual forms randomly sampled from the semantic space of the knowledge.
However, existing LLMs are notorious for being sensitive to prompt, thereby undermining the reliability of such evaluations \citep{Ji_2023,maharana2023exposing,chang2023language,chen2023say}.
Consequently, current studies on model evaluation are reasonably considered to be insufficiently robust.

\begin{figure}
\centering
    \includegraphics[width=0.9\linewidth]{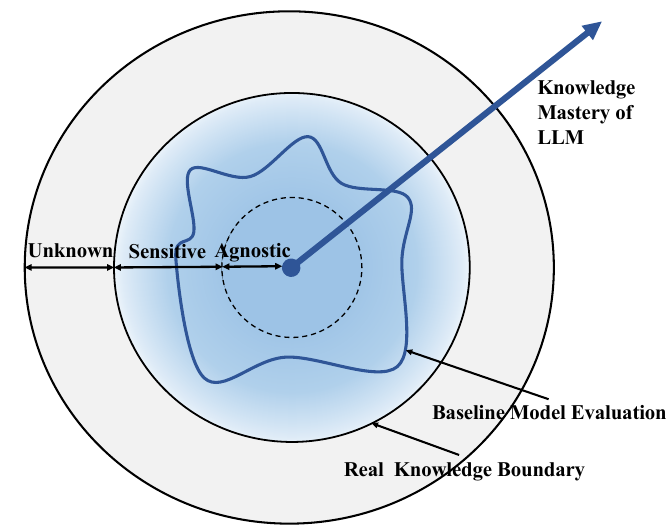}
    \caption{Illustration of three classes of knowledge based on the model's mastery of knowledge in different textual forms. Existing evaluation methods suffer from sensitivity to input prompt. Therefore, the knowledge ability depicted by these methods is irregularly shaped. We propose to evaluate the knowledge capacity with a knowledge boundary containing both Prompt-Agnostic Knowledge and Prompt-Sensitive Knowledge.}
\label{fig:kbing}
\end{figure}

As shown in Figure \ref{fig:kbing}, from the perspective of the model's mastery of the textual form of knowledge, knowledge can be divided into three classes: 1) \textbf{Prompt-Agnostic Knowledge} that can be correctly answered for any textual form; 2) \textbf{Prompt-Sensitive Knowledge} that is sensitive to the form of the prompt fed into the model; 3) \textbf{Unanswerable Knowledge} that is unable to be answered by the model, regardless of the prompt employed.
The majority of previous research on model evaluation ignored the presence of Prompt-Sensitive Knowledge, resorting to oversimplified binary evaluations, classifying the model's knowledge mastery merely as true or false.
\citet{dong2023statistical} attempts to assess LLM through diverse paraphrases, yet these evaluations remain confined to limited textual forms of knowledge.
We give strict definitions of three types of knowledge in Section \ref{sec:knowledge}.

In this paper, we aim to reduce the contingency when evaluating LLMs. 
Different from previous paradigms of LLM evaluation, we attempt to explore the Unanswerable Knowledge of the model to be evaluated, thereby illuminating the knowledge boundaries of LLMs. 
How can we find Unanswerable knowledge for the model? It is obvious that trying all prompts for the knowledge to query the model is too resource-intensive. Therefore, we choose to make efforts to search for the optimal prompt. 
We formalize optimal prompt searching as a discrete optimization problem: given some question paraphrases, we search for a prompt to maximize the probability of generating the correct answer.
We propose the Projected Gradient Descent method with Constraints (PGDC), a new algorithm that updates prompt with gradient descent and implements proximal projection to search discrete prompts.
To ensure that the optimized prompt has the same semantics as the original prompt, we introduce semantic loss, which is a measure of the distance between the semantic representations of the optimized prompt and the original prompt.

Experimental results demonstrate that our proposed PGDC can outperform baselines in depicting knowledge boundaries.
In addition, results on counterfactual datasets demonstrate that our approach is reasonable and robust. 
Human evaluation also reveals that our optimized prompts generally have the same semantics as the original questions.
Moreover, we delineate models' knowledge boundaries in different domains using PGDC to evaluate LLMs. 
The size of the model's domain knowledge boundaries is strongly associated with the performance of downstream tasks in the domain.
The optimal prompts also have some patterns that can give some inspiration for designing prompts when using corresponding LLMs.

In summary, our contributions are: 
(1) We propose a new evaluation paradigm for benchmarking knowledge boundaries to compare models' capabilities, which can reduce the randomness in current evaluations.
(2) We design PGDC, a projected gradient descent method with constraints, to optimize prompts and obtain knowledge boundaries of LLMs which achieves the best results on four datasets.
(3) We evaluate five models using knowledge boundaries and obtain some valuable findings.

Our code and data are released to facilitate future research\footnote{\href{https://github.com/pkulcwmzx/knowledge-boundary}{https://github.com/pkulcwmzx/knowledge boundary}}.

\begin{figure*}
    \centering
    \includegraphics[width=0.9\linewidth]{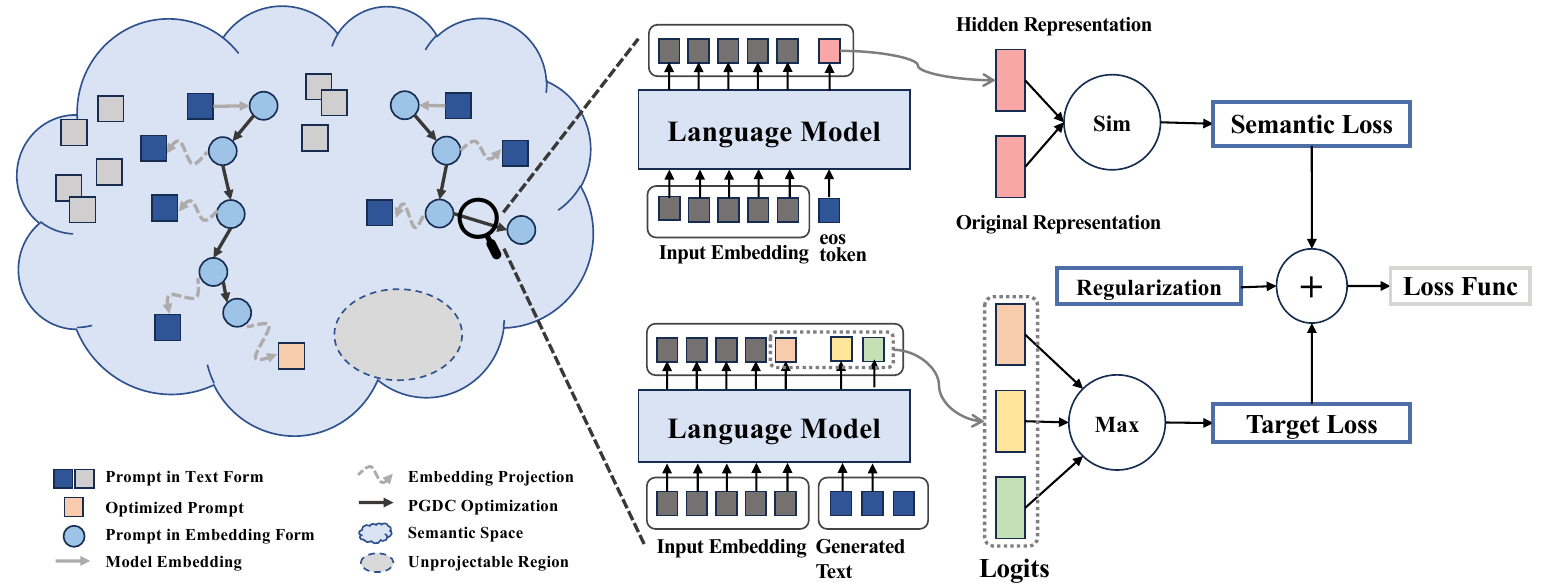}
    \caption{An illustration of our PGDC method, projected gradient descent method with constraints. The left part of the image shows the overall framework of our method: we start from a few labeled prompts, perform gradient descent with the target answer as the optimization goal and try to project the embedding into text form, while ensuring that the whole search process is in the same semantic space of the expression of the target knowledge. The right side of the image shows how our loss function is calculated at each step of gradient descent.}
    \label{fig:pgdc}
\end{figure*}

\section{Preliminaries}
\subsection{Knowledge Boundary Definition}
\label{sec:knowledge}

First, we provide a strict definition of the three types of knowledge in LLMs, Prompt-Agnostic Knowledge, Prompt-Sensitive Knowledge and Unanswerable Knowledge.

Formally, let $k$ denote a piece of knowledge, and let $\Theta$ represent an LM. We assess whether $\Theta$ "possesses" knowledge of k by calculating $P(a_k^i|q_k^i, \Theta)$, where $q_k^i \in Q_k, a_k^i \in A_k$ represents a pair of input-output tokens to verify the knowledge $k$. 

For the universe of all conceivable knowledge $U$, and a threshold $\epsilon$ within the range $(0.5, 1]$, the three types of knowledge for model $\Theta$ are defined as follows:
\begin{list}{\labelitemi}{\leftmargin=1em}
    \item Prompt-Agnostic Knowledge: $K^A=\{k\in U| P(a_k^i|q_k^i, \Theta)>\epsilon, \forall q_k^i \in Q_k, \forall a_k^i \in A_k\}$
    \item Unanswerable Knowledge: $K^U=\{k\in U| P(a_k^i|q_k^i, \Theta)< \epsilon, \forall q_k^i \in Q_k, \forall a_k^i \in A_k\}$
    \item Prompt-Sensitive Knowledge: $K^S = \{k\in U| k \notin K^U  \land k \notin K^A\}$
\end{list}
In short, for a piece of knowledge $k$, if the model $\Theta$  is able to answer the question about $k$ correctly under at least one expression, $k$ is inside the knowledge boundary of $\Theta$. If $\Theta$  is unable to answer the question about $k$ correctly under any expressions, $k$ is outside the knowledge boundary of the model.

\subsection{Knowledge Boundary Requirements}
We attempt to calculate the knowledge boundary of LLM by automatically constructing the optimal prompt.
As various methods for prompt engineering have been proposed to obtain better prompt as query \citep{dong2022survey, wei2023chainofthought}, not all of them are suitable for calculating the knowledge boundary of LLM.
In this section, we propose four basic requirements for the algorithm applied to the calculation of knowledge boundaries: Universality, Truthfulness, Robustness and Optimality.
\paragraph{Universality}
When searching for an optimal prompt, the method should work for most current LLMs, regardless of its size and architecture.

\paragraph{Truthfulness}
The constructed prompt should share the same semantics as the original question, and not be allowed to change subject or relation.

\paragraph{Robustness}
When searching for the optimal prompt for a piece of knowledge, the effectiveness of the method should be relevant to the knowledge capacity of LLM.
In other words, the algorithm should tend not to find appropriate prompt for unanswerable knowledge.

\paragraph{Optimality}
The algorithm should search for as much prompt-sensitive knowledge in the LLM as possible.

\subsection{Problem Formulation}
In this section, we give a formal problem formulation of searching for the optimal prompt.
For a given piece of knowledge, assume we have an LLM that models next-token probability $P(x_i|x_1, x_2, ..., x_{i-1})$ with an input sequence $(x_1, x_2, ..., x_{i-1})$.
The piece of knowledge is expressed in various textual forms to construct a QA set of multiple questions and answers.
Different questions in the QA set are paraphrases, while answers are aliases.
We believe that if the model is able to answer one of the questions correctly, it is possible for the model to "know" this piece of knowledge.
Therefore, if the model is able to generate one of the correct answers with prompt semantically similar to one of the questions, we consider the knowledge within its knowledge boundary.

To illustrate the problem, we start from the simpler case with only one question $Q = \{q_1, q_2, ..., q_n\}$ and one answer $A= \{a_1, a_2, ..., a_m\}$.
Prompt $X$ is initialized with $Q$ and optimized to maximize the probability of generating $A$ while remaining semantically similar to $Q$.
We formalize optimal prompt searching as the problem:
\begin{align}
\centering
    \min_{X}{\Phi(X) = L(X, A)+\lambda R(X, Q)}, 
\end{align}
where $L(\cdot)$ denotes the loss function to penalize unsuccessful generation. 
$R(\cdot)$ indicates the semantic distance between the optimized prompt and the initial prompt while $\lambda$ is the penalty factor.

\section{Method}
To obtain a better knowledge boundary for LLM, our effort is directed toward identifying the optimal prompt within the semantic space.
As illustrated in Figure \ref{fig:pgdc}, PGDC optimizes prompt in the neighbor semantic space of the original question.
The prompt in text form is first mapped to prompt in embedding form as continuous text embedding.
During PGDC optimization, the text embedding is updated through gradient descent with direction of the loss function $\Phi(\cdot)$.
After each update, if the text embedding is close to a discrete prompt, it is projected to the discrete prompt through embedding projection.
To avoid the text embedding from entering unprojectable region where there are no close discrete prompts to project, we introduce a regularization to force the embedding not to enter these regions. 
After multiple iterations of the update, we get the final optimized prompt.

\subsection{PGDC Optimization}
In PGDC algorithm, we do not specify the positions of the answer in our LLM-generated output, which relaxes constraints in the model output and leaves space for the LLM to generate reasoning process and do inference.
Therefore, we define the target loss of generating a specific answer $A$ with a slicing window method:
\begin{align}
\centering
    L = \min_{j<=t-k_i+1}{-\log P(O_{j:j+k_i}=A)},
\end{align}
where $O = \{o_1, o_2, ..., o_t\}$ denotes the output of the LLM given $X$ as the input. In this way, PGDC automatically searches for the target position in the model output and optimizes the probability of generating answer.
When there exist multiple answers in the answer set $A^*$, we optimize the answer with the highest probability to be generated:
\begin{align}
\centering
    L = \min_{A \in A^*}\min_{j<=t-k_i+1}{-\log P(O_{j:j+k_i}=A)}.
\end{align}
We separately optimize prompts with PGDC if there are multiple paraphrases of questions in the piece of knowledge.

Since PGDC optimizes prompts in the continuous embedding space while text space is discrete, it is hard for methods of automatically searching for prompts to constrain semantic information (\citet{shin-etal-2020-autoprompt}; \citet{2303.04381}).
To combat the challenge, we introduce a semantic constraint to the loss function, which is defined as:
\begin{align}
\centering
    R(X, Q) = ||h(X) - h(Q)||_2, 
\end{align}
where $h(\cdot)$ is the hidden representation of prompt and $||\cdot||_2$ denotes the L2 distance between two items.
As illustrated in Figure \ref{fig:pgdc}, the hidden representation is obtained with the last hidden layer output of the LLM given the concatenation of prompt and a <eos> character.

As the optimization process is implemented in the continuous space, it is necessary to project the embedding into discrete tokens.
The optimized embedding obtained might enter the unprojectable region shown in Figure \ref{fig:pgdc}, which makes the projection hard. 
Therefore, we add a regularization in the loss function to punish prompt embedding far from discrete tokens:
\begin{align}
\centering
    \delta(X) = \Sigma_{i=1}^{N}{\min_{v \in \mathcal{V}}{||\hat x_i-Wv||_2}}, 
    \label{eq:loss}
\end{align}
where $\mathcal{V}$ denotes the vocabulary of LLM and $W$ is the projection from vocab to embedding space.
The lowercase letters such as $x$ represent tokens while $\hat x$ represents its embedding.
$N$ in Equation \ref{eq:loss} denotes the length of the prompt $X$ which varies in different iterations.

In general, the final loss function of PGDC is formulated as: 
\begin{align}
\centering
    \Phi(X) = L(X, A)+\lambda_1 R(X, Q) + \lambda_2 \delta(X).
\end{align}

\subsection{Proximal Projection}
Instead of projecting the prompt into text space after the overall optimization \cite{guo-etal-2021-gradient} or conducting projection after each iteration \cite{cheng2020seq2sick}, PGDC achieves flexible transformation of embedding space to text space with a threshold of the vector distance.
Formally, the transformation can be written as:
\begin{align}
\centering
    \hat x_i = 
    \left\{
    \begin{array}{lr}
        Wv, & \min_{v \in \mathcal{V}}{||\hat x_i-Wv||_2} < c \\
        \hat x_i, & \min_{v \in \mathcal{V}}{||\hat x_i-Wv||_2} >= c, \\
    \end{array}
    \right.
\end{align}
where $c$ represents the threshold of the $L2$ distance.
As illustrated in Figure \ref{fig:pgdc}, the dashed line shows the proximal projection process.

\subsection{Algorithm Summary}
In general, PGDC iteratively optimizes prompt in the embedding space with gradient descent to minimize the loss function in Equation~\ref{eq:loss} and do proximal projection after each iteration.
A detailed pseudocode is shown in Appendix \ref{sec:alg}.

\section{Experiments}
In this section, we perform comparisons between our method and baseline methods which are commonly used in model evaluation on common knowledge benchmarks and unanswerable knowledge benchmarks. We also conduct a manual evaluation to check whether the semantics of the prompts we obtained are consistent with the original question.

\begin{table*}[t]
\begin{center}
\resizebox{2.0\columnwidth}{!}{
\begin{tabular}{c|c|ccccccc}
\toprule
\multirow{2}{*}{\bf Dataset} & \multirow{2}{*}{\bf Model} & \multicolumn{7}{c}{\bf Method} \\
\cmidrule{3-9}
& & zero & few & dis & P-zero & P-few & P-dis & \bf PGDC(ours) \\
\midrule \midrule
\multirowcell{4}{\textsc{PaRaRel}$\uparrow$} 
& LLaMA2 & 34.43\% & 58.23\% & 17.96\% & 54.78\% & 66.95\% & 44.16\% & \bf 71.36\% \\
& Vicuna & 34.19\% & 59.56\% & 8.40\% & 54.97\% & \bf 69.69\% & 23.06\% &  69.63\% \\
& GPT-J &  23.23\% & 44.84\% & 2.40\% & 40.78\% & 54.06\% & 7.25\% & \bf 55.95\% \\
& GPT-2 & 9.27\% & 12.61\% & 3.13\% & 18.18\% & 20.46\% & 9.01\% & \bf 47.68\% \\
\midrule 
\multirowcell{4}{KAssess$\uparrow$} 
& LLaMA2 & 23.69\% & 32.03\% & 6.73\% & 50.00\% & 50.75\% & 24.73\% & \bf 69.84\% \\
& Vicuna & 23.21\% & 33.37\% & 9.90\% & 51.15\% & 53.66\% & 38.20\% & \bf 57.63\% \\
& GPT-J &  15.95\% & 20.47\% & 12.67\% & 40.23\% & 38.20\% & 2.26\% & \bf 48.62\% \\
& GPT-2 & 4.03\% & 3.64\% & 2.46\% & 13.66\% & 11.44\% & 15.75\% & \bf 24.71\% \\
\midrule \midrule
\multirowcell{4}{CFACT$\downarrow$} 
& LLaMA2 &\bf 1.32\% & 4.56\% & 13.88\% & 3.30\% & 9.31\% & 36.46\% &  3.41\% \\
& Vicuna &\bf 1.40\% & 3.08\% & 4.95\% & 3.36\% & 6.91\% & 14.28\% & 3.50\% \\
& GPT-J & \bf 1.39\% & 3.18\% & 2.30\% & 3.75\% & 6.12\% & 6.74\% & 4.82\% \\
& GPT-2 &\bf 1.10\% & 1.77\% & 3.28\% & 3.00\% & 3.92\% & 9.32\% & 2.81\% \\
\midrule
\multirowcell{4}{\textsc{AlCuna}$\downarrow$} 
& LLaMA2 &\bf 0.00\% & 0.63\% & 30.48\% & \bf 0.00\% & 0.63\% & 30.48\% & \bf 0.00\% \\
& Vicuna &\bf 0.00\% & 0.80\% & 0.90\% & \bf 0.00\% & 0.80\% & 0.90\% & \bf 0.00\% \\
& GPT-J & \bf 0.00\% & 0.08\% & 0.72\% & \bf 0.00\% & 0.08\% & 0.72\% & \bf 0.00\% \\
& GPT-2 & \bf 0.00\% & 0.30\% & 2.10\% & \bf 0.00\% & 0.30\% & 2.10\% & \bf 0.00\% \\
\bottomrule
\end{tabular}
}
\caption{The success rate of constructing prompts to elicit specific knowledge on four Datasets. We conduct experiments on four different LLMs to illustrate the performance of our proposed PGDC. Dataset \textsc{PaRaRel} and KAssess provide true knowledge to characterize the ability of different methods to obtain knowledge boundary while the pieces of knowledge in dataset CFACT and \textsc{AlCuna} are fake which shows the robustness of PGDC. 
}
\label{tab:result}
\end{center}
\end{table*}

\subsection{Datasets and Models}
\paragraph{Common Knowledge Benchmarks}
In order to evaluate the performance of different methods on common knowledge, we choose to use KAssess \citep{dong2023statistical} and \textsc{PaRaRel} \citep{elazar-etal-2021-measuring}  for our evaluation.
Both of them consist of knowledge tuples and hand-curated prompt templates, where all subjects, relations, and objects exist as entities in WikiData.

\paragraph{Unanswerable Knowledge Benchmarks}
To test whether our optimized prompts leak answers or induce hallucinations that cause LLMs to answer knowledge that they could not answer originally, we perform evaluations on two counterfactual datasets, COUNTERFACT (denoted as CFACT) \citep{meng2022locating} and \textsc{AlCuna} \citep{yin-etal-2023-alcuna}.
CFACT contains 20K counterfactual knowledge records with a diverse set of subjects, relations, and linguistic variations.
\textsc{AlCuna} is a biological dataset for evaluating the ability of the model in face of new knowledge.

The above datasets have multiple expressions for each knowledge query, except for \textsc{AlCuna}.

\paragraph{Models}
Our experiments use GPT-2 (774M) \citep{radford2019language}, GPT-J (6B) \citep{gpt-j}, LLaMA2 (7B) \citep{touvron2023llama}, and Vicuna (7B) \citep{vicuna2023}.

\subsection{Baseline Methods}
There are several common methods of assessing the model's mastery of knowledge, and we use the following as baselines:
\paragraph{Zero-Shot (zero)} 
Zero shot prompting is the simplest and most common approach used in previous evaluation work. 
We directly query models using questions from benchmarks.
\paragraph{Few-Shots (few)}
Few shots prompting is commonly used to enhance model performance by utilizing the contextual learning capabilities of LLMs. We retrieve similar knowledge in the dataset as exemplars to feed to the model. 
\paragraph{Discriminator (dis)}
We can also use the judgment question format to assess whether a model knows one piece of knowledge. 
So we provide LLM with one knowledge statement and let it determine whether this statement is correct or incorrect. 

Since there are several paraphrases for each knowledge query in the benchmarks, for each above baseline, we will use two different metrics to simulate previous work on model evaluation: 
1) For each knowledge query, we will randomly select one of its expressions for evaluation.
2) For each knowledge query, as long as one of its paraphrases can be answered correctly, the knowledge is considered to be inside the boundaries, and the baseline method using this metric is denoted as P-baseline.

A more detailed description of the dataset as well as the implementation of the baseline methods, are shown in Appendix \ref{app:datamodel}.
The hyperparameter settings for the PGDC are shown in Appendix \ref{app:hyper}.

\subsection{Results and Analysis}
Table \ref{tab:result} summarizes the experimental results on four different LLMs. 
Our proposed PGDC achieves the highest performance on common knowledge benchmarks on almost all LLMs.
The results indicate that the knowledge boundary found by our method is more comprehensive than baseline methods, which shows the \textit{\textbf{Optimality}} and \textit{\textbf{Universality}} of PGDC. 
The experimental results on unanswerable knowledge benchmarks including CFACT and \textsc{AlCuna} reflect the \textit{\textbf{Robustness}} of different prompt methods.
PGDC only slightly raises the amount of unanswerable knowledge over zero-shot baseline, which shows that our proposed method will only introduce relatively limited fake knowledge and meets the \textit{\textbf{Robustness}} requirement. PGDC considers knowledge to be within the boundaries if any of its paraphrases can be answered correctly for each knowledge query and shows comparative results with P-zero while outperforming P-few and P-dis.
Moreover, the prompts generated by PGDC are generally consistent semantically with the original questions (as shown in Section \ref{sec:he}).
Therefore, PGDC meets all four fundamental criteria (Universality, Truthfulness, Robustness, and Optimality)  in calculating knowledge boundaries.

In addition, we can observe that:
\paragraph{Evaluating LLMs with a fixed question or limited paraphrases as the query is not reliable and comprehensive}
According to the zero and P-zero results of the \textsc{PaRaRel} and the KAssess dataset, we can see that different queries yield different results, and different prompting methods may result in different inter-system rankings. This suggests that assessing LLMs using a predetermined question or restricted paraphrases as the query lacks reliability and comprehensiveness. Evaluating LLMs with a fixed question or limited paraphrases may lead to the selection of suboptimal LLMs for practical applications, demonstrating the necessity of optimizing prompt design to seek more realistic knowledge boundaries.

\paragraph{Discrimination format is much less reliable than cloze-style format}
P-dis has a similar proportion of responses as true on common and unanswerable knowledge benchmarks, which correlates with the model's preference for true. This observation aligns with previous findings by  \citet{wu2023style,wang2023large}.

\paragraph{Different models prefer different prompts}
Since traditional model evaluation methods use fixed queries, the model's preference for prompt affects the score.
The difference between P-zero and zero then reflects the fact that the model is sensitive to prompt. Even for queries with the same meaning, different ways of asking can produce different results.
GPT-2 also acquires a fair amount of knowledge, but is overly sensitive and thus scores lower on traditional assessment methods.

\paragraph{Manual design of a good prompt is difficult}
Few shots prompting induces more knowledge than zero-shots. However, it is difficult to verify how to select good examples and whether a good enough prompt has been designed.

PGDC, on the other hand, uses cloze-style problem and automatically searches for the optimal prompt for different models, so it is a much better approach to model evaluation.


\begin{figure}
\centering
    \includegraphics[width=0.8\linewidth]{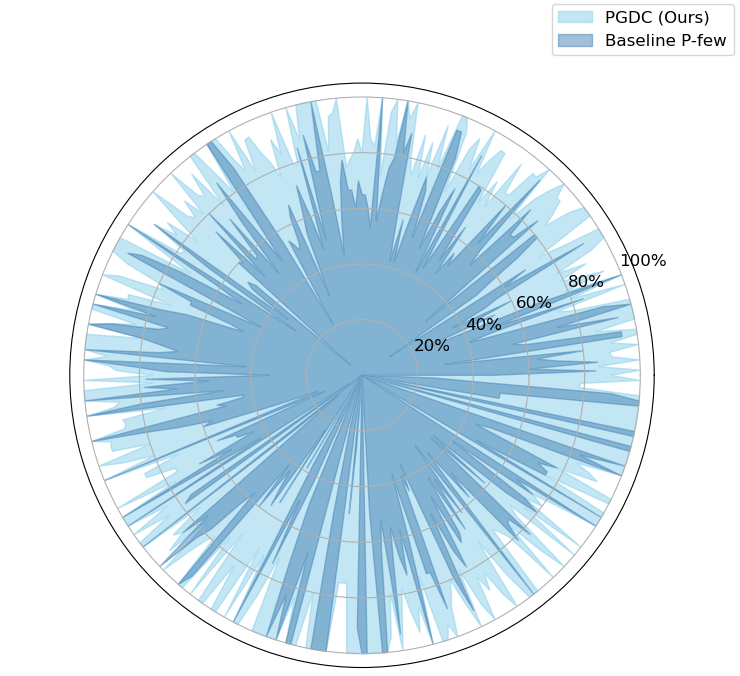}
    \caption{Knowledge boundaries of PGDC and baseline method P-few on KAssess using LLaMA2 model.}
    \label{fig:staAna}
\end{figure}

We also analyze the knowledge detected by PGDC as well as the knowledge detected by baselines on KAssess. 
We categorize relations according to KAssess, and analyze the accuracy of PGDC and baseline methods on various relation categories. The results of PGDC and the strong baseline method P-few on the strong LLaMA2 model are shown in Figure \ref{fig:staAna}, while the coverage results of other baseline methods are presented in the Appendix \ref{app:couverage}.
We find that the knowledge boundaries we obtained can almost cover baselines.
\begin{figure}
\centering
    \includegraphics[width=0.9\linewidth]{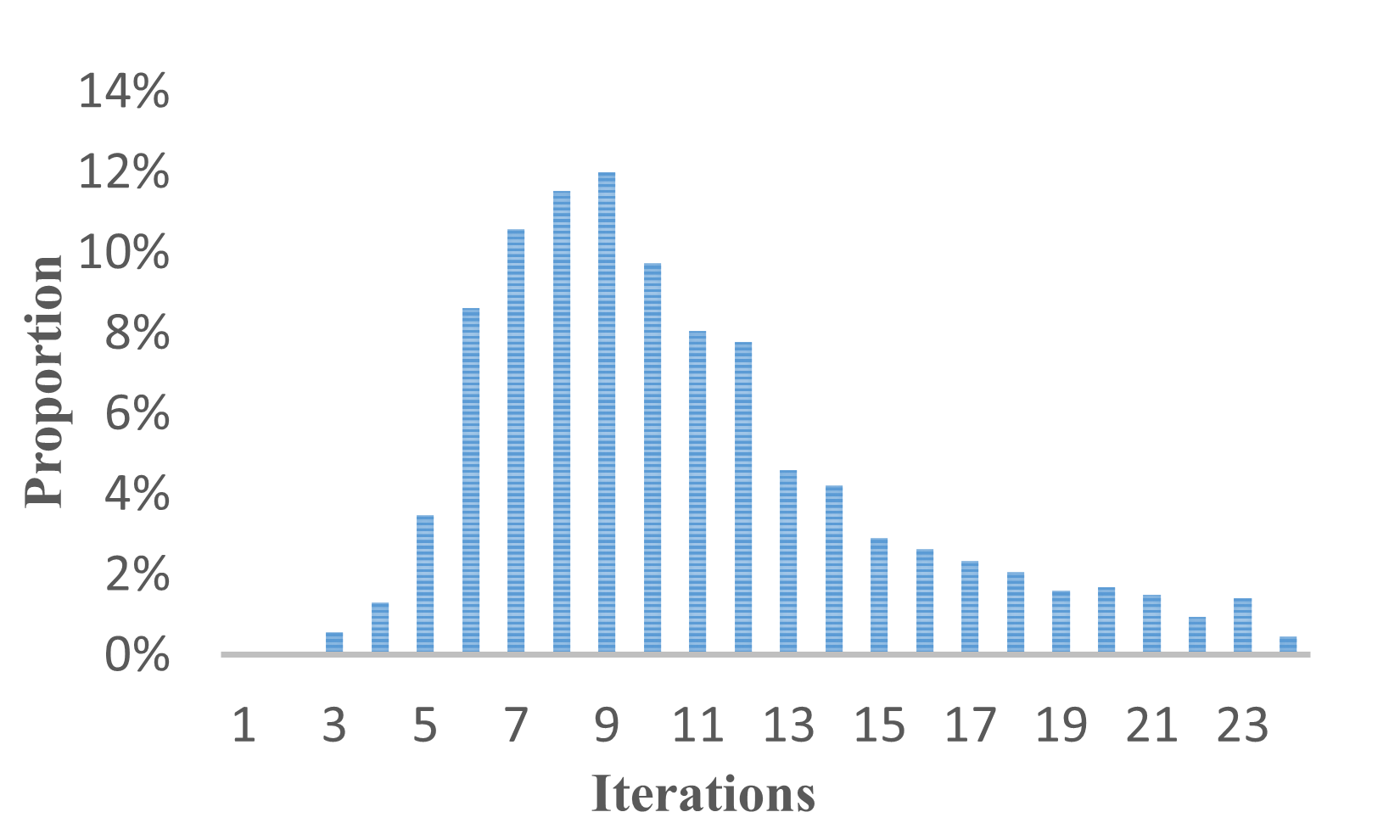}
    \caption{Iterations on KAssess to find the optimized prompt using PGDC with LLaMA2 model.}
    \label{fig:distri}
\end{figure}
Moreover, we also record the iterations on KAssess to find the optimized prompt using PGDC in Figure \ref{fig:distri}.
We observe that PGDC can find the optimal prompt for the majority of queries within 15 iterations.


\begin{table*}[t!]
    \centering
    \begin{tabular}{p{15cm}}
    \toprule[2pt]
    \textcolor{blue}{[Original Prompt]} The associated item of source code is; \quad
    \textcolor{green}{[Answer]} program \\
    \textcolor{blue}{[PGDC Prompt]} The early item of source code is a simple; \quad
    \textcolor{orange}{[Optimal Paraphrase]} \\
    \hline
    \textcolor{blue}{[Original Prompt]} Isatis tinctoria is a source of; \quad
    \textcolor{green}{[Answer]} indigo \\
    \textcolor{blue}{[PGDC Prompt]} Isatis tinctoria, a source of the natural dye; \quad
    \textcolor{orange}{[Reasoning and Inference]} \\
    \hline
    \textcolor{blue}{[Original Prompt]} The host country of Australian Capital Territory is; \quad
    \textcolor{green}{[Answer]} Australia \\
    \textcolor{blue}{[PGDC Prompt]} <s> host country of Australian Capital Territory is; \quad
    \textcolor{orange}{[Format and Stop Words]} \\
    \bottomrule[2pt]
    \end{tabular}
    \caption{Cases that PGDC successfully updates the prompt. We summarize the advantages of PGDC into three aspects : 1) Optimal Paraphrase; 2) Reasoning and Inference; 3) Format and Stop Words.}
    \label{tab:example}
\end{table*}

\begin{table}[]
\resizebox{\columnwidth}{!}{
\begin{tabular}{ccccc}
\toprule
Dataset    & \multicolumn{4}{c}{CFACT$\downarrow$}  \\ \midrule
Model      & GPT-2        & GPT-J       & LLaMA2        & Vicuna       \\ \hline
Autoprompt & 92.38\%      &  85.67\%    &    88.35\%    &  33.09\%           \\
PGDC       & \bf 2.81\%   & \bf 4.82\%  & \bf3.41\%     &  \bf3.50\%           \\ \bottomrule
\end{tabular}
}

\caption{Comparison of PGDC and AutoPrompt on CFACT dataset.}
\label{tab:autop}
\end{table}

\subsection{Comparison with Prompt Optimization Method}
Since our method is a prompt optimization type of method, we conduct experiments to compare the robustness of PGDC and Autoprompt \citep{shin-etal-2020-autoprompt}, a representative method of prompt optimization.
Autoprompt is a Hotflip-based algorithm \cite{ebrahimi-etal-2018-hotflip} in optimizing prompt, which employs several trigger tokens to elicit the target output.
The exact experimental setup is shown in Appendix \ref{app:autoprompt}.

As shown in Table \ref{tab:autop}, we can find that Autoprompt induces the model to output target answers on counterfactual datasets in a large percentage. 
This result suggests that Autoprompt is more similar to an adversarial attack algorithm that is committed to getting the target answer, while PGDC optimizes the prompt within the semantic constraint.

\subsection{Semantic Preservation Evaluation}\label{sec:he}
In order to check whether the prompt obtained by PGDC is semantically consistent with the original questions (\textit{\textbf{Truthfulness}}), we perform a manual evaluation.
We randomly select 200 samples from the \textsc{PaRaRel} dataset.
Specifically, we enlisted three college students who hold English qualification certificates. Initially, they were given an evaluation guideline, which is detailed in Appendix \ref{sec:anno_guideline}. Each evaluator underwent a training process to improve their comprehension of the annotation procedure. 
Prior to annotation, we administered a qualification test comprising 10 samples; only annotators who passed this test were deemed qualified and permitted to proceed with annotation. 

The human evaluation results show that the semantic preservation rates of GPT-2, GPT-J, LLaMA2, and Vicuna are respectively 80.5\%, 85.1\%, 83.3\%, and 86.2\%. This indicates that the prompts generated by PGDC are generally semantic consistent with the original questions, which demonstrates the general Truthfulness of PGDC. More details about the human evaluation are shown in Appendix \ref{sec:anno_guideline}.


\subsection{Case Study}
To understand how PGDC steers question prompts to generate desired answers, we manually study cases in which PGDC successfully updates the prompt.
We summarize the advantages of PGDC into three aspects and provide cases in Table \ref{tab:example}: 
1) Finding the optimal paraphrase of the original prompt. 
Due to human resource constraints, it is impossible to enumerate all paraphrases of the original question.
PGDC automatically searches for the optimal paraphrase that elicits correct answers.
2) Leaving space for LLM to reason and infer.
PGDC allows LLMs to generate some tokens to assist their reasoning and inference to achieve the answer.
3) Changing the format and stop words in the original prompt.
Some special tokens and stop words vary in different LLMs, which can be hard for humans to detect.
PGDC is able to optimize format and stop words on the basis of gradient.

\section{Assessments of LLMs}
The above experiments have demonstrated the effectiveness of PGDC in detecting knowledge and the reasonableness of the obtained optimal prompt. In this section, we apply PGDC on MMLU  \citep{hendrycks2021measuring} to evaluate LLMs.

\subsection{Experimental Settings}
We evaluate GPT-2 (774M), GPT-J (6B), LLaMA2 (7B), Vicuna (7B) and Mistral (7B) \citep{jiang2023mistral} from the perspective of 30 refined domain knowledge using MMLU\footnote{MMLU covers 57 subjects.
To fit the theme of our paper, here we have selected 30 topics related to knowledge, dropping topics such as computation and reasoning, for analysis.
}.

To be consistent with our approach, we modify the questions in MMLU from choice questions to a cloze format, which yields more reliable and stable assessment results.
Following previous work \citep{anil2023palm,touvron2023llama1}, we categorize the questions in MMLU into six types of topics: natural sciences, medical and biological sciences, computer science and logic, social sciences, humanities, and others.
More details of the experiment are shown in Appendix \ref{app:mmluexp}.

\begin{figure}
\centering
    \includegraphics[width=0.9\linewidth]{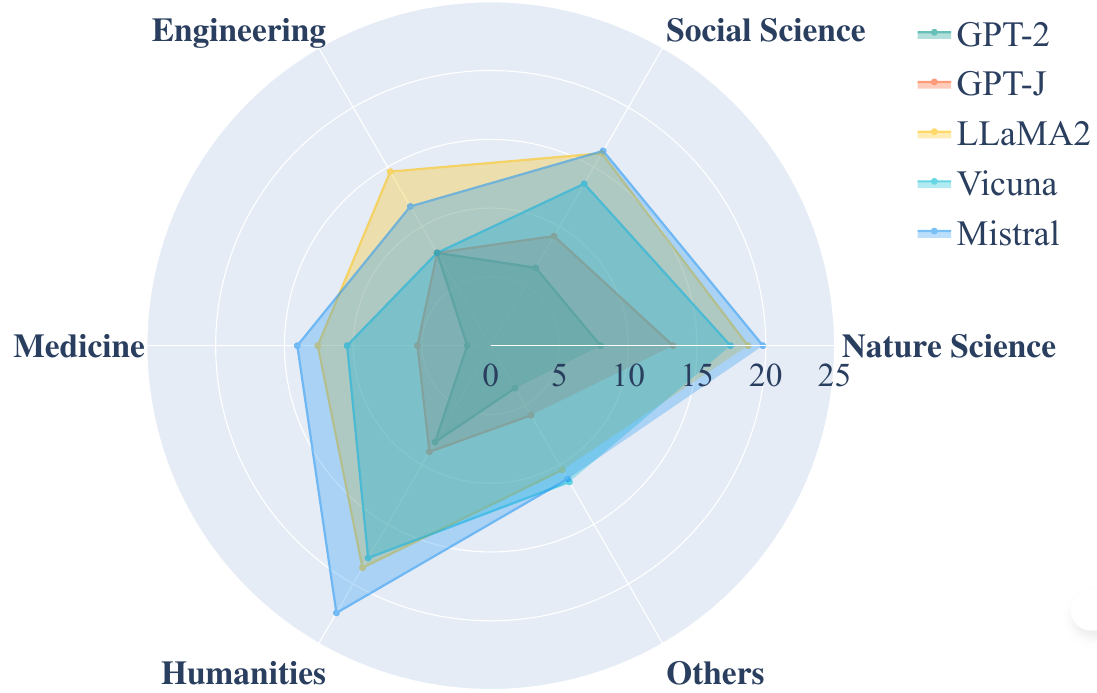}
    \caption{Knowledge boundaries of different domains of models on MMLU.}
    \label{fig:mmlu}
\end{figure}

\subsection{Results}
The results\footnote{For display purposes, our radar graph ranges 1-25\%.} of our model evaluation on each of the broad categories are demonstrated in Figure \ref{fig:mmlu}, and more detailed scores are shown in Appendix \ref{app:mmlures}.

We can find that Mistral has the largest knowledge boundaries overall. LLaMA2 exceeds the other models by a lot in the engineering domain. 
It may be because LLaMA2 uses a lot more new labeled code data for training.
Vicuna performs not far behind LLaMA2 on other topics. GPT-2 has very small knowledge boundaries and performs poorly in the more specialized medical domain.
By identifying more reliable knowledge boundaries, we help select the appropriate LLM in practice.

However, it is worth noting that these scores are quite low (around 20 points). This is due to the fact that we use more difficult cloze-style questions for reliability. It also reflects the fact that the results obtained from the choice-style benchmark may be too high. The problems in MMLU are relatively specialized, and the current general-purpose models do not have a lot of knowledge in the relevant areas.

\section{Discussion of Randomness in Model Evaluation}
The inherent randomness in model evaluation presents significant challenges to the reliability of evaluation results and the process of model selection.
This randomness can be seen in several key areas:

\paragraph{Benchmark Question Selection}
Randomness is inherent in the selection of benchmark questions for testing, as the objective is to choose representative questions that accurately demonstrate the model's abilities in specific areas or domains \citep{guo2023close,zhong2023agieval}.
\paragraph{Prompt Expression}
Formulating the selected questions into prompts introduces additional randomness, including aspects such as the phrasing of questions, system instructions, the order of options, and other details. 
Numerous studies have demonstrated that models are sensitive to variations in prompts \citep{Ji_2023,maharana2023exposing,chang2023language,chen2023say}.
\paragraph{Decoding Settings}
The choice of decoding parameters (e.g., temperature) during evaluation can introduce randomness into the model's output, potentially raising concerns about fairness.
\paragraph{Output Evaluation}
The evaluation of the model's generated content also involves randomness. 
Some studies use exact matching \citep{yin-etal-2023-alcuna}, while others employ GPT-4 for evaluation \citep{vicuna2023}; however, neither method is flawless.

Given the various sources of randomness in model evaluation, developing reliable, stable, and informative evaluation methods remains a significant challenge and warrants further research. 
To address this concern, we introduce the concept of the "knowledge boundary," which employs optimized prompts to test the upper limits of a model's capabilities, thereby minimizing the randomness associated with prompt selection.

\section{Related Work}
\paragraph{Model Evaluation}
Several benchmarks have been proposed to evaluate Large Language Models (LLMs) on human exams like college entrance and law school admission tests \citep{suzgun2022challenging, srivastava2023imitation, choudhary2023complex, zhong2023agieval}. 
In terms of knowledge assessment, LAMA \citep{petroni-etal-2019-language} evaluates whether models can correctly predict masked object entities in a cloze-style prompt.
Some studies \citep{onoe2021creak,mallen2023trust,arodi2023kitmus,yu2023kola} focus on measuring LLMs' understanding and mastery of world knowledge.
These benchmarks do not take into account that LLMs are sensitive to different prompts.
Some works \citep{elazar-etal-2021-measuring,dong2023statistical} focus on estimating and measuring the consistency of LLMs given diverse prompts.
All previous studies on model evaluation use fixed prompts, and our work pioneers prompt optimization for evaluating LLMs' knowledge boundaries.

\paragraph{Prompt Optimization}
Due to the sensitivity of language models to prompts, better prompts can help achieve higher performance in specific tasks (\citet{deng-etal-2022-prompt}; \citet{wei2023chainofthought}; \citet{yang2023large}).
Prompt engineering like in-context learning greatly improves the performance of prompt methods \cite{dong2022survey}.
Another related line of work attempts to formalize prompt searching as a discrete optimization task to achieve better performance in specific tasks \cite{shin-etal-2020-autoprompt}.
Some studies adopt Hotflip-based algorithms \cite{ebrahimi-etal-2018-hotflip} to automatically construct prompts (\citet{wallace-etal-2019-universal}; \citet{shin-etal-2020-autoprompt}; \citet{2303.04381};).
In addition, several work tries to optimize prompts in continuous embedding space with Gumbel-softmax trick \cite{guo-etal-2021-gradient} and projection (\citet{cheng2020seq2sick}; \citet{wen2023hard}).

\section{Conclusion}
The sensitivity of LLMs to prompt leads to the unreliability of the results obtained from traditional model evaluation works that use fixed queries to evaluate the model. 
To address this problem, we propose semantics-preserving prompt optimization methods, PGDC, to find the knowledge boundaries of models for model evaluation.
Our experiments demonstrate shortcomings of previous model evaluation methods and the fact that the prompt we find is superior to the fixed prompt. At the same time, the prompt found by our method maintains the original semantics and does not induce knowledge that is not captured by the model, which outperforms previous prompt optimization efforts.
Moreover, we conduct experiments exploring the boundaries of the model's different domain knowledge and compare and analyze the LLM's capabilities.

\section*{Limitations}
According to our definition, one can say that a model knows this knowledge when it can answer the corresponding question with the optimal prompt. 
In this paper, we only aim to find the Unanswerable Knowledge of the model as the knowledge boundary. 
In fact, for Prompt-sensitive knowledge, the model's sensitivity also reflects the model's mastery of it (the knowledge in the color gradient in Figure \ref{fig:kbing}). At this stage we would like to have a clear boundary, so we have not considered this part for now. But exploring this part of knowledge is an interesting and important future work.

Additionally, despite our efforts to ensure the truthfulness of optimized prompts, there remains a small probability that the semantics of the prompt may change, especially for weaker language models such as GPT-2.

\section*{Ethics Statement}
A potential negative impact of our approach is that malicious attackers could use our method to attack public large pre-trained language models, leading to fake knowledge generation. 
As pre-trained language models advance in many tasks, addressing safety concerns becomes increasingly necessary and imperative.
Analyses in our paper can help enhance the evaluation of pre-trained language models.

\section*{Acknowledgements}
This work was supported by Beijing Science and Technology Program (Z231100007423011), National Key R\&D Program of China (2021YFF0901502), National Science Foundation of China (No. 62161160339) and Key Laboratory of Science, Technology and Standard in Press Industry (Key Laboratory of Intelligent Press Media Technology). We would like to thank Zhidong Jia, Yezhen Chen, Baizhou Huang, Junzhe Zhang and members of the group for their valuable feedback and discussions. We appreciate the anonymous reviewers for their helpful comments. Xiaojun Wan is the corresponding author.


\bibliography{anthology,custom}
\bibliographystyle{acl_natbib}
\newpage
\appendix
\section{Datasets and Baselines}
\label{app:datamodel}
\subsection{Datasets}
We conduct comparative experiments between our method and the baseline on four datasets. The datasets are described and detailed below:
\paragraph{KAssess} KAssess \citep{dong2023statistical} is a large-scale assessment suite with 994,123 entities, 600 relations, and their text aliases which are obtained from T-REx knowledge graph \citep{elsahar-etal-2018-rex}.
KAssess constructs multiple paraphrased templates for each relation. 
In total, there are 3,488 templates for 600 relations, with an average of 5.82 paraphrased templates per relation.
\paragraph{\textsc{PaRaRel}} \textsc{PaRaRel} \citep{elazar-etal-2021-measuring} is also a manually curated resource that provides patterns—short textual prompts—that are paraphrases of one another, with 328 paraphrases describing 38 binary relations.

\paragraph{COUNTERFACT} COUNTERFACT \citep{meng2022locating} is an  evaluation dataset for evaluating counterfactual edits in language models which contains 21,919 records with a diverse
set of subjects, relations, and linguistic variations.
We use its target knowledge as counterfactual knowledge to query LLMs.

\paragraph{\textsc{AlCuna}} \textsc{AlCuna} \citep{yin-etal-2023-alcuna} is used for evaluating the ability of LLMs in the face of new knowledge which consists of a total of 84351 questions about 3554
artificial entities.
We only select the cloze-style portion of the questions to be used for the experiments.

\begin{table*}[]
\centering
\begin{tabular}{m{1.0cm}<{\centering}m{3.5cm}<{\centering}m{10.5cm}}
\toprule
                             
\multirow{3}{*}{KAssess}     & Zero-Shot     & 1. 10,000 metres record is held by {[}Kenenisa Bekele{]}                                                                                                                                              \\
                             & Few-Shots     & 2. Pole vault record is held by Fabiana Murer.\textbackslash{}t 800 metres record is held by David Rudisha\textbackslash{}t 10,000 metres record is held by {[}Kenenisa Bekele{]}                     \\
                             & Discriminator & 3. Check whether the following statement is correct: 10,000 metres record is held by Kenenisa Bekele. The statement is (True/False): {[}True{]}                                                       \\ \midrule
\multirow{3}{*}{\textsc{PaRaRel}}     & Zero-Shot     & 1. The mother tongue of Go Hyeon-jeong is {[}Korean{]}                                                                                                                                                \\
                             & Few-Shots     & 2. The mother tongue of Michel Denisot is French. \textbackslash{}t The mother tongue of Thomas Joannes Stieltjes is Dutch. \textbackslash{}t The mother tongue of Go Hyeon-jeong is {[}Korean{]}     \\
                             & Discriminator & 3. Check whether the following statement is correct: The mother tongue of Go Hyeon-jeong is Korean. The statement is (True/False): {[}True{]}                                                         \\ \midrule
\multirow{3}{*}{CFACT} & Zero-Shot     & 1. IBM Connections is created by {[}Adobe{]}                                                                                                                                                          \\
                             & Few-Shots     & 2. Windows Embedded CE 6.0 is created by IBM. \textbackslash{}t Sandy Bridge was a product of Apple. \textbackslash{}t IBM Connections is created by {[}Adobe{]}                                      \\
                             & Discriminator & 3. Check whether the following statement is correct: IBM Connections is created by Adobe. The statement is (True/False): {[}True{]}                                                                   \\ \midrule
\multirow{3}{*}{\textsc{AlCuna}}      & Zero-Shot     & 1. What's the body length of Leuciaiaivea? {[}8.1 cm{]}                                                                                                                                               \\
                             & Few-Shots     & 2. What's the body length of Octopus perralis? 100.0 cm. \textbackslash{}t What's the body length of Sepia bidabilis? 17.0 cm. \textbackslash{}t What's the body length of Leuciaiaivea? {[}8.1 cm{]} \\
                             & Discriminator & 3. Check whether the following statement is correct: What's the body length of Leuciaiaivea? 8.1 cm. The statement is (True/False): {[}True{]}                                                        \\ \bottomrule
\end{tabular}
\caption{Demonstration of baselines. Answers are in '{[}{]}'. Each query has multiple textual expressions and each answer has multiple aliases. The number of examples for few-shots in our experiments is 4. Due to space constraints, we do not show these comprehensively.}
\label{tab:inputexample}
\end{table*}

\subsection{Baselines}
We slightly adapted the dataset to the characteristics of the generative model, and examples of the inputs are shown in Table \ref{tab:inputexample}.

\section{Hyperparameter setting for PGDC}
\label{app:hyper}

Hyperparameter settings are shown in Table \ref{tab:configuration}.

\begin{table} [] 
\centering
\begin{tabular}{cc} 
\hline
\multicolumn{2}{c}{Hyperparameter} \\ 
\hline
Learning rate & 1e-2 \\
Optimizer & Adam  \\
Scheduler & ExponentialLR \\
Schedule Step & 5 \\
Iteration Rounds & 25 \\
$\lambda_2$ & 0.01 \\
\hline

\end{tabular}
\\
\caption{\label{tab:configuration}
Hyperparameter settings of PGDC.
}
\end{table}

\section{Pseudocode for our algorithm}
\label{sec:alg}
We provide pseudocode for ASRA is in Algorithm \ref{algo}.

\begin{algorithm}
    \caption{\textbf{PGDC} Algorithm}
    \label{algo}
    Input: LLM $\theta$, Embedding Table $E^{|\mathcal{V}|}$, Input Question $q = \{q_1, q_2, ..., q_n\}$, Answer $a$, Loss Function $\Phi$, Optimization Step $T$, Learning Rate $\gamma$, Projection Ceil c
    \begin{algorithmic}[1]
        \State $p \gets E^{|\mathcal{V}|}[q]$
        \For {$i = 1, 2, ... T$}
            \State Generate $z = \{z_1, z_2, ..., z_m\}$ with $p$ as an input into $\theta$
            \State $L \gets \Phi(z)$
            \State $p \gets p - \gamma\nabla_{p}{L}$
            \For{$p_j \in p$}
                \State $t \gets \arg\min_{k \le |\mathcal{V}|} {||p_j - v_k||_2}$
                \If {$||p_j - v_t||_2 \le c$}
                    \State $p_j \gets v_t$
                \EndIf
            \EndFor
        \EndFor
        \State \Return $p$
    \end{algorithmic} 
\end{algorithm}

\section{Coverage Analysis}
\label{app:couverage}
\begin{figure*}[htbp]
	\centering
 \subfloat[Knowledge boundaries of PGDC and baseline P-few.]{\includegraphics[width=.45\linewidth]{figure/rader.png}}\hspace{5pt}
	\subfloat[Knowledge boundaries of PGDC and baseline few.]{\includegraphics[width=.45\linewidth]{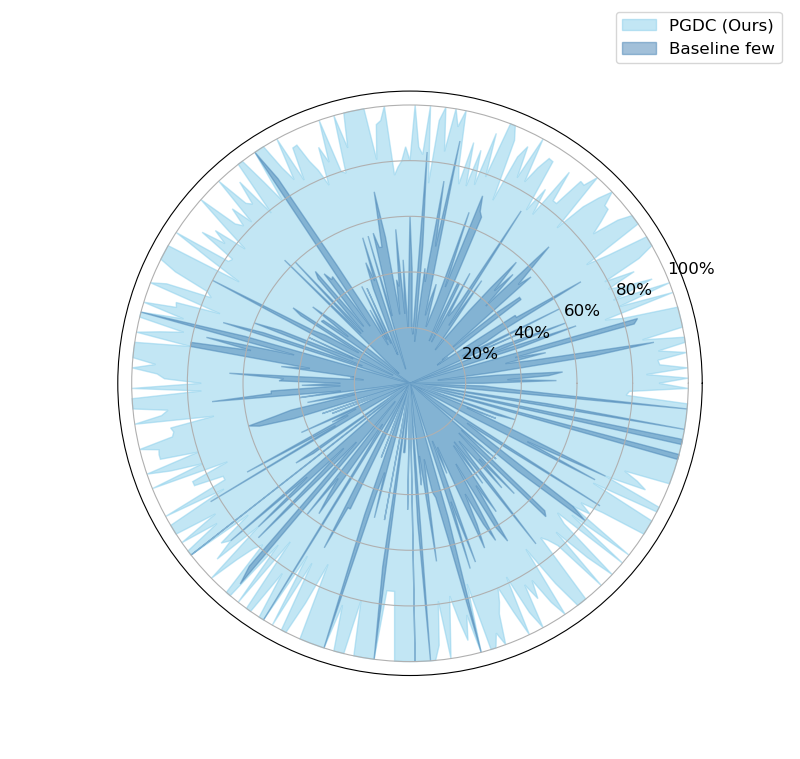}}\\
	\subfloat[Knowledge boundaries of PGDC and baseline P-zero.]{\includegraphics[width=.45\linewidth]{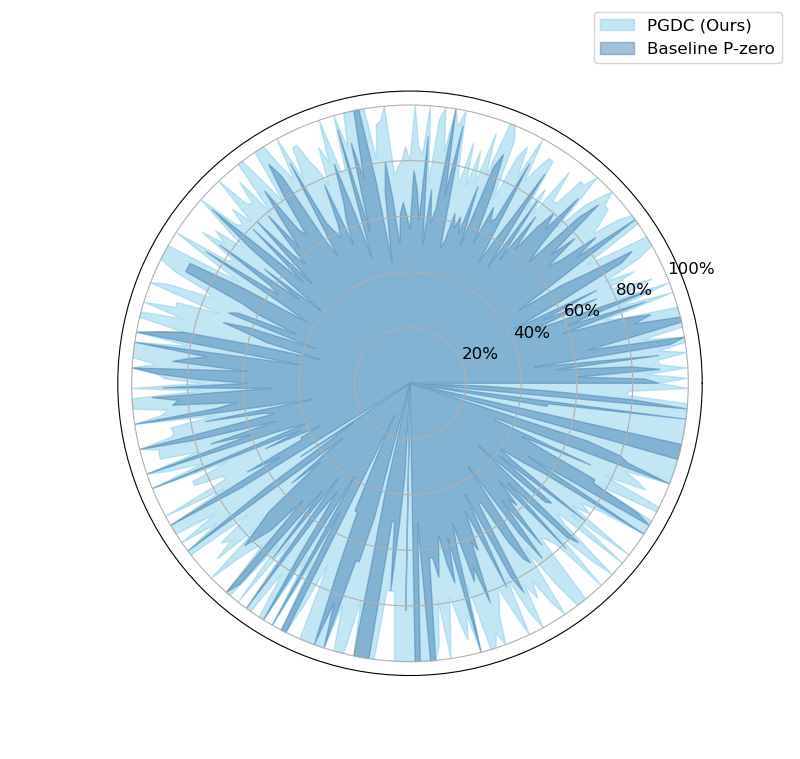}}\hspace{5pt}
	\subfloat[Knowledge boundaries of PGDC and baseline zero.]{\includegraphics[width=.45\linewidth]{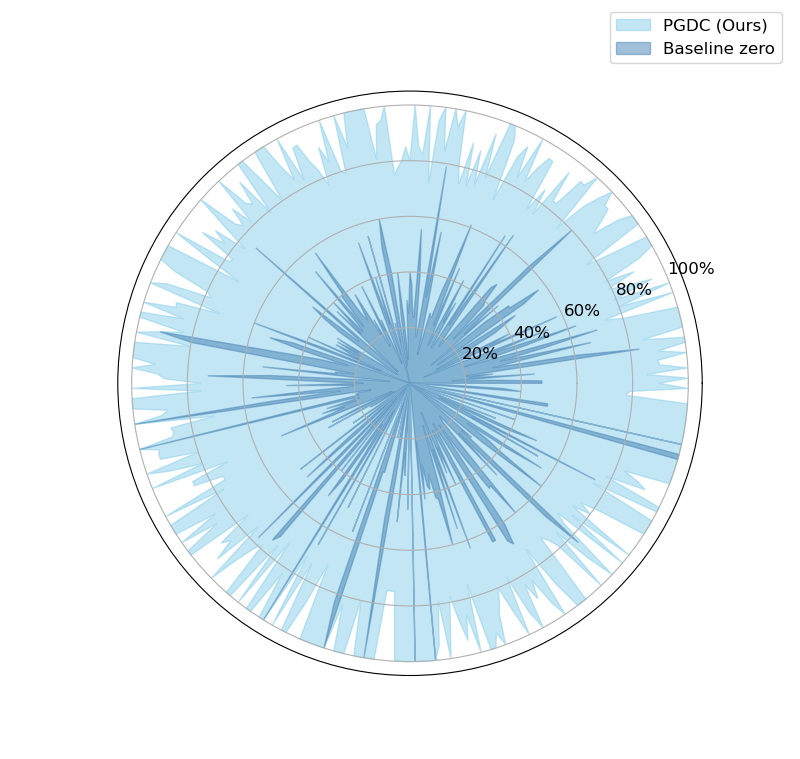}}\\
	\subfloat[Knowledge boundaries of PGDC and baseline P-dis.]{\includegraphics[width=.45\linewidth]{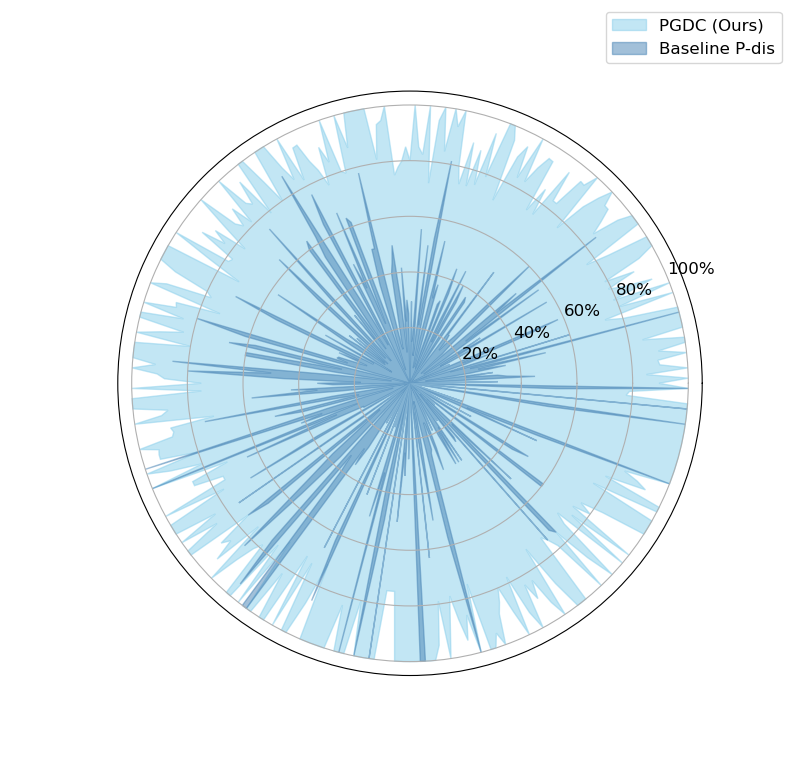}}\hspace{5pt}
	\subfloat[Knowledge boundaries of PGDC and baseline dis.]{\includegraphics[width=.45\linewidth]{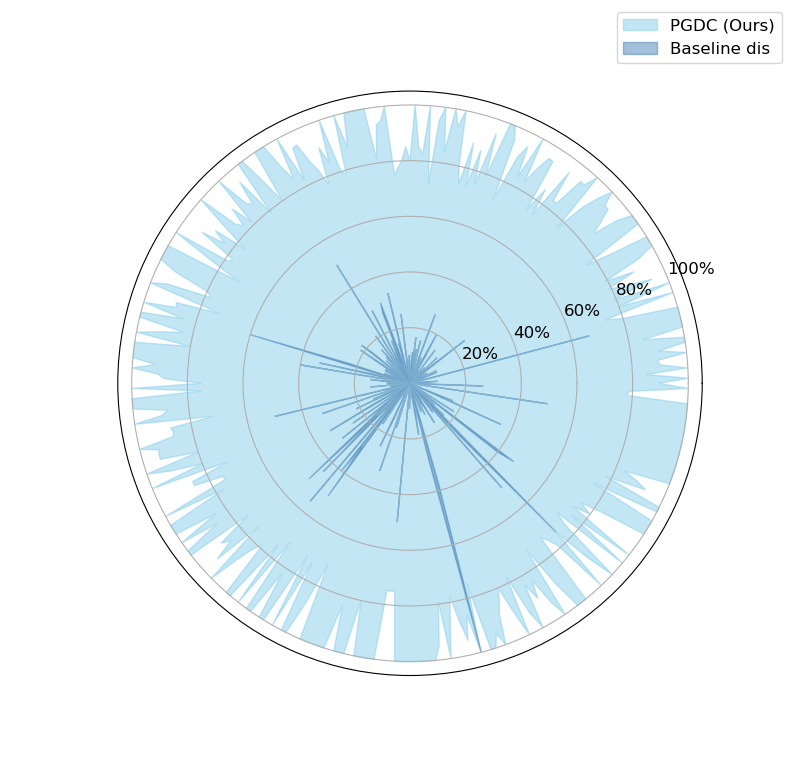}}\hspace{5pt}
	\caption{Knowledge boundaries of the proposed PGDC and baseline methods on KAssess using LLaMA2 model.}
 \label{fig:rader-sub}
\end{figure*}

We also examined the knowledge identified by PGDC, as well as the knowledge identified by baseline methods based on KAssess.
The result of PGDC and the baseline methods P-few, few, P-zero, zero, P-dis and dis on the strong LLaMA2 model are shown in Figure \ref{fig:rader-sub}.
Our analysis reveals that the knowledge boundaries we derived can effectively encompass those of the baseline methods.

\section{Autoprompt for Model Evaluation}
\label{app:autoprompt}
We implement Autoprompt by extending the question with five trigger tokens initialized with the last token in original prompt.
The trigger tokens are updated for three rounds according to the algorithm described in \citet{shin-etal-2020-autoprompt}.

\section{Human Evaluation}
\label{sec:anno_guideline}
\begin{table}[t]\small
\centering
\setlength{\extrarowheight}{0pt}
\addtolength{\extrarowheight}{\aboverulesep}
\addtolength{\extrarowheight}{\belowrulesep}
\setlength{\aboverulesep}{0pt}
\setlength{\belowrulesep}{0pt}
\resizebox{\linewidth}{!}{
\begin{tabular}{|p{7cm}|} 
\hline
\multicolumn{1}{|c|}{{\large\textbf{ Human Evaluation Guideline}}}                                            \\ 
\hline
\textbf{\large\textcolor{red}{Task Overview}}                                     \\
Thank you for participating in this task! We are currently working on a project focused on benchmarking knowledge boundary for Large Language Model (LLM). You will be randomly presented with two texts. Your task is to determine whether the semantics of two texts are consistent, which means judging whether the two texts possess consistent semantics and can both be used to inquire about the same knowledge. Note that the texts do not need to be fluent or grammatically correct, and the presence of non-disruptive gibberish that does not affect the recognition of semantics is allowed. If the semantics of the two texts are consistent, label 1; if not, label 0. Please maintain high quality in your annotations.\\  \hline  
\textbf{\large\textcolor{red}{Emphasis and Caution}}                                     \\

\textbf{Support and Reference}: If you encounter any confusion regarding professional knowledge or context while performing this task, please feel free to reach out to us for clarification.You may also refer to Wikipedia or other reliable sources to gain further understanding.

\textbf{Feedback Mechanism}: You can directly submit your queries, concerns, or suggestions to us.
\\ \hline
\end{tabular}}
\caption{Human evaluation guideline.}\label{tab:guidelines}
\end{table}
We provide our human evaluation guideline furnished to participants for manually evaluating the semantic preservation task, as presented in Table \ref{tab:guidelines}. 

We recruited three college students, all possessing College English Test-6 certificates, demonstrating fluency in English. 
We first distribute the evaluation guidelines to the evaluators. Subsequently, we conduct training sessions for the evaluators, explaining the evaluation guidelines to ensure a better understanding of the task requirements and addressing any questions or concerns they may have.
Before commencing formal annotation tasks, we administered a qualification test. Ten samples were randomly selected. These samples were evaluated by the participants, and subsequently, we assessed the accuracy of each annotator's evaluations. A higher accuracy score reflects a more consistent understanding of our guidelines. Evaluators who achieved at least 90 \% accuracy were deemed qualified to proceed with the evaluation task.
We employed Fleiss's Kappa statistic \citep{fleiss1981measurement} to assess the agreement between the three annotators, yielding a score of 0.64.



\section{Model Evaluation on MMLU}
\label{app:mmlu}
\subsection{Experimental Settings}
\label{app:mmluexp}
Based on the conclusion of our earlier experiments that cloze-style questions are more reliable, we converted the choice-style questions in MMLU to a cloze format.
We remove the other options and only keep the contents of the correct option as the answer to the cloze question.

\begin{table*}[]
\begin{tabular}{llccccc}
\toprule
Broader Subject                 & Subject in MMLU                               & GPT-2 & GPT-J & LLaMA2 & Vicuna & Mistral \\ \midrule
\multirow{7}{*}{Nature Science} & astronomy\_test                               & 5.08  & 6.78  & 6.78   & 10.17  &    15.25     \\
                                & college\_biology\_test                        & 2.56  & 10.26 & 14.10  & 10.26  &    16.67     \\
                                & college\_chemistry\_test                      & 1.89  & 1.89  & 11.32  & 9.43   &    5.67     \\
                                & conceptual\_physics\_test                     & 14.47 & 19.30 & 23.25  & 23.25  &      23.25   \\
                                & high\_school\_physics\_test                   & 0.00  & 0.00  & 1.72   & 1.72   &    1.72     \\
                                & high\_school\_biology\_test                   & 1.41  & 8.45  & 14.79  & 11.97  &   15.49      \\
                                & high\_school\_chemistry\_test                 & 3.06  & 4.08  & 7.14   & 6.12   &    8.16     \\ \midrule
\multirow{9}{*}{Social Science} & high\_school\_government\_politics & 10.34 & 12.64 & 18.39  & 16.09  &    14.94     \\
                                & high\_school\_macroeconomics\_test            & 0.93  & 1.87  & 3.74   & 4.67   &     5,61    \\
                                & high\_school\_microeconomics\_test            & 2.50  & 2.50  & 5.00   & 4.17   &    5.00     \\
                                & management\_test                              & 1.08  & 0.00  & 5.38   & 5.38   &    2.15     \\
                                & professional\_accounting\_test                & 1.67  & 0.00  & 1.67   & 1.67   &     0.00    \\
                                & sociology\_test                               & 0.00  & 3.49  & 6.98   & 3.49   &    5.81    \\
                                & us\_foreign\_policy\_test                     & 5.08  & 10.17 & 10.17  & 6.78   &    10.17     \\
                                & world\_religions\_test                        & 3.61  & 4.21  & 19.88  & 15.06  &    18.07     \\
                                & high\_school\_psychology\_test                & 10.08 & 14.47 & 19.90  & 17.05  &    22.22     \\ \midrule
\multirow{2}{*}{Engineering}    & electrical\_engineering\_test                 & 6.02  & 5.26  & 10.53  & 6.02   &    6.77     \\
                                & college\_computer\_science\_test              & 0.00  & 4.76  & 4.76   & 0.00   &     14.29    \\ \midrule
\multirow{6}{*}{Medicine}       & clinical\_knowledge\_test                     & 0.00  & 2.68  & 4.70   & 2.68   &     6.71    \\
                                & college\_medicine\_test                       & 2.86  & 2.86  & 10.00  & 2.86   &    8.57     \\
                                & medical\_genetics\_test                       & 0.00  & 4.11  & 9.59   & 13.70  &    9.59     \\
                                & nutrition\_test                               & 2.53  & 3.16  & 10.13  & 7.59   &  8,87       \\
                                & virology\_test                                & 0.79  & 2.38  & 3.97   & 3.97   &  6.35       \\
                                & anatomy\_test                                 & 1.23  & 9.88  & 20.99  & 19.75  &    25.93     \\ \midrule
\multirow{3}{*}{Humanities}     & global\_facts\_test                           & 1.14  & 3.41  & 7.95   & 9.09   &     12.50    \\
                                & moral\_disputes\_test                         & 3.52  & 2.01  & 7.04   & 4.52   &    6.03     \\
                                & miscellaneous\_test                           & 10.30 & 11.61 & 23.37  & 22.79  &    28.45     \\ \midrule
\multirow{3}{*}{Others}         & high\_school\_geography\_test                 & 4.85  & 8.48  & 16.36  & 12.73  &     13.94    \\
                                & logical\_fallacies\_test                      & 0.00  & 0.98  & 2.94   & 3.92   &    3.92     \\
                                & human\_aging\_test                            & 29.27 & 3.90  & 5.37   & 9.76   &    8.29     \\ \midrule
\end{tabular}
\caption{Our categorization of subjects in MMLU and detailed scores.}
\label{tab:mmlutopic}
\end{table*}

Since the topic of our paper is about knowledge and some of the questions in MMLU are about computation and reasoning, we filter them out
The remaining 30 subjects are grouped into six larger subjects, as shown in Table \ref{tab:mmlutopic}.

The PGDC method in this experiment uses the same hyperparameters as in Appendix \ref{app:hyper}.

\subsection{Detailed Results}
\label{app:mmlures}
In the main article we report the results in the broad categories, and the results in each subcategory are shown in Table \ref{tab:mmlutopic}.

\end{document}